\documentclass[conference]{IEEEtran}
\IEEEoverridecommandlockouts

\usepackage{amsmath,amssymb,amsfonts}
\usepackage{algorithmic}
\usepackage{graphicx}
\usepackage{booktabs}
\usepackage{textcomp}
\usepackage{xcolor}
\usepackage{url}
\usepackage{hyperref}
\usepackage{subcaption}   
\usepackage{caption}      

\usepackage[left]{lineno}
\begin{document}

\title{Synthetic-to-Real Object Detection using YOLOv11 and Domain Randomization Strategies}

\author{\IEEEauthorblockN{Luisa Torquato Niño}
\IEEEauthorblockA{
MIT-KIT, Germany}
\and
\IEEEauthorblockN{Hamza A. A. Gardi}
\IEEEauthorblockA{
IIIT at ETIT- KIT, Germany}
}

\maketitle

\begin{abstract}
This paper addresses the synthetic-to-real domain gap in object detection, focusing on training a YOLOv11 model to detect a specific object (a soup can) using only synthetic data and domain randomization strategies. The methodology involves extensive experimentation with data augmentation, dataset composition, and model scaling. While synthetic validation metrics were consistently high, they proved to be poor predictors of real-world performance. Consequently, models were also evaluated qualitatively, through visual inspection of predictions, and quantitatively, on a manually labeled real-world test set, to guide development. Final mAP@50 scores were provided by the official Kaggle competition. Key findings indicate that increasing synthetic dataset diversity, specifically by including varied perspectives and complex backgrounds, combined with carefully tuned data augmentation, were crucial in bridging the domain gap. The best performing configuration, a YOLOv11l model trained on an expanded and diverse dataset, achieved a final mAP@50 of 0.910 on the competition's hidden test set. This result demonstrates the potential of a synthetic-only training approach while also highlighting the remaining challenges in fully capturing real-world variability.
\end{abstract}


\section{Introduction}
The development of object detection models for real-world applications is often restricted by the high cost and effort of collecting and labeling large datasets. Synthetic data offers a cost-effective, scalable, and unbiased way to create labeled training data, and is increasingly recognized for its potential to address significant real-world challenges in various domains \cite{goyal}. However, the domain gap between synthetic training data and real-world test data creates a challenge to applying this technique to real-world applications, leading to poor generalization performance \cite{vanherle2022analysistrainingobjectdetection}. 

In this context, a Kaggle contest \cite{kaggle} was created to train a YOLO model for soup can detection using only synthetic data generated with Falcon's Duality AI Simulator. This paper details a project created for this contest that focuses on solving the synthetic-to-real domain gap using only synthetic data and domain randomization strategies with the ultimate goal of achieving a high Mean Average Precision at IoU 0.5 (mAP@50) score. 

Although the use of YOLOv8 pre-trained models was initially proposed by Falcon, an independent decision was made to change to YOLOv11 from Ultralytics \cite{yolo11_ultralytics}, as benchmarking comparisons (see Figure \ref{fig:benchmarkyolo}) indicated superior performance of YOLOv11 in terms of accuracy, speed and efficiency. Moreover, data augmentation strategies were explored to improve the model's ability to generalize from the synthetic domain to unseen real-world images from the test set.

This paper is divided into a brief overview of related work, a description of the methodology, dataset, model, and experimental setup. Finally, the results are presented and discussed, ending the paper with remarks on potential future work.

\section{Related work}
\begin{figure}[htb!]
    \begin{center}
    \caption{Benchmarking YOLOv11 Against Previous Versions.}
    \includegraphics[width=8cm]{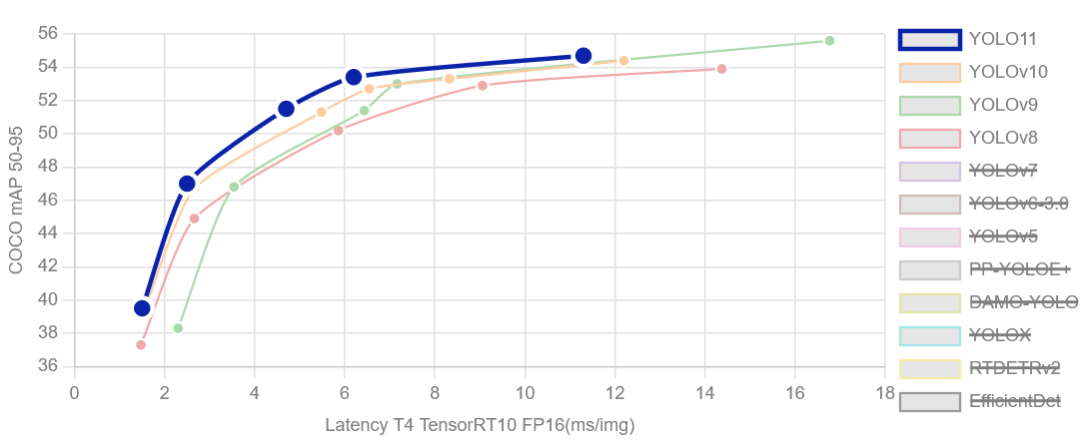}
    \footnotesize{\\Source: Ultralytics \cite{yolo11_ultralytics}.}
    \label{fig:benchmarkyolo}
	\end{center}
\end{figure}

Object detection is a fundamental computer vision task with numerous applications. Recent advances have been significantly driven by deep learning models, particularly the YOLO (You Only Look Once) series \cite{yolo_original}, which frame detection as a regression problem to offer real-time performance. Following a suggestion from the Kaggle competition to use YOLO, a benchmarking comparison of performance metrics across different YOLO versions \ref{fig:benchmarkyolo} showed that YOLOv11 achieves higher accuracy, which was the main goal of the Kaggle challenge.  

Training robust object detectors typically requires large and labeled datasets such as COCO \cite{cocodataset} or ImageNet \cite{imagenet}. The difficulty in obtaining such datasets for specific objects or scenarios has led to an increased interest in the use of synthetic data \cite{bauer2024comprehensiveexplorationsyntheticdata} \cite{Mumuni_2024} \cite{vanherle2022analysistrainingobjectdetection}. However, training on synthetic data and testing on real-world data creates a sim-to-real domain gap, which is the challenge imposed by this Kaggle competition. 

Approaches to bridge this gap are diverse and can be broadly categorized. One major approach is Domain Adaptation, which aims to align the feature distributions between the synthetic and real domains. This is often achieved through adversarial training, in which a domain discriminator network is trained to distinguish between synthetic and real features, while the main model is trained to fool this discriminator, thus learning domain-invariant features \cite{ganin2015unsuperviseddomainadaptationbackpropagation}. Other methods focus on image-to-image translation techniques, using Generative Adversarial Networks (GANs) such as CycleGAN to transform synthetic images to look more realistic before they are used for training \cite{8237506}.

In contrast to these methods, which often require complex training pipelines or architectural modifications, Domain Randomization offers a more direct approach \cite{tzeng2017adversarialdiscriminativedomainadaptation} \cite{tobin2017domainrandomizationtransferringdeep} \cite{chen2022understandingdomainrandomizationsimtoreal}, aiming to train models on synthetic data with sufficient variability, such that the real-world appears as just another variation. This paper focuses on the latter approach, specifically leveraging data augmentation as a form of domain randomization. This method is computationally efficient and does not require access to unlabeled real-world data during training, making it suitable for challenges like the one addressed in the Kaggle competition.

This project leverages the strong foundation of the YOLO architecture and explores practical data augmentation strategies as a form of domain randomization to improve synthetic-to-real transfer for a specific object detection task.

\section{Methodology} \label{methodology}
The methodology employed in this project focused on training a YOLOv11 \cite{yolo11_ultralytics} model on a synthetic dataset and evaluating different approaches to maximize its performance on unseen real-world images for the specific task of soup can detection.

\subsection{Object Detection Model: YOLOv11}
The YOLOv11 model was initialized with weights pre-trained on the COCO dataset to leverage features learned from a large-scale real-world dataset \cite{cocodataset}. Developed by Ultralytics \cite{yolo11_ultralytics} and released in 2024, YOLOv11 builds upon the foundation of YOLOv8 by aiming to further optimize the balance between speed, accuracy and computational efficiency. 

Key architectural changes in YOLOv11, as summarized by Khanam et al. \cite{khanam2024yolov11overviewkeyarchitectural} in comparison to YOLOv8, include: 
\begin{itemize}
    \item C3K2 Block: An evolution of the C3/CSP (Cross-Stage Partial) that enhances feature extraction efficiency.
    \item SPPF (Spatial Pyramid Pooling - Fast):  An implementation optimized for speed while effectively capturing context by pooling features at multiple scales.
    \item C2PSA (Convolutional block with Parallel Spatial Attention): The integration of attention mechanisms to enable the network to focus on the most relevant regions within feature maps. 
    \item Refined Backbone/Neck: The underlying feature extractor and fusion network likely incorporate these new blocks to improve information flow and the quality of feature maps passed to the detection head.
\end{itemize}

\begin{figure}[htb!]
    \begin{center}
    \caption{YOLOv11 features a refined architecture with advanced C3k2 blocks, SPPF, and C2PSA modules that significantly enhance multi-scale feature extraction and spatial attention for improved detection accuracy.}
    \includegraphics[width=8cm]{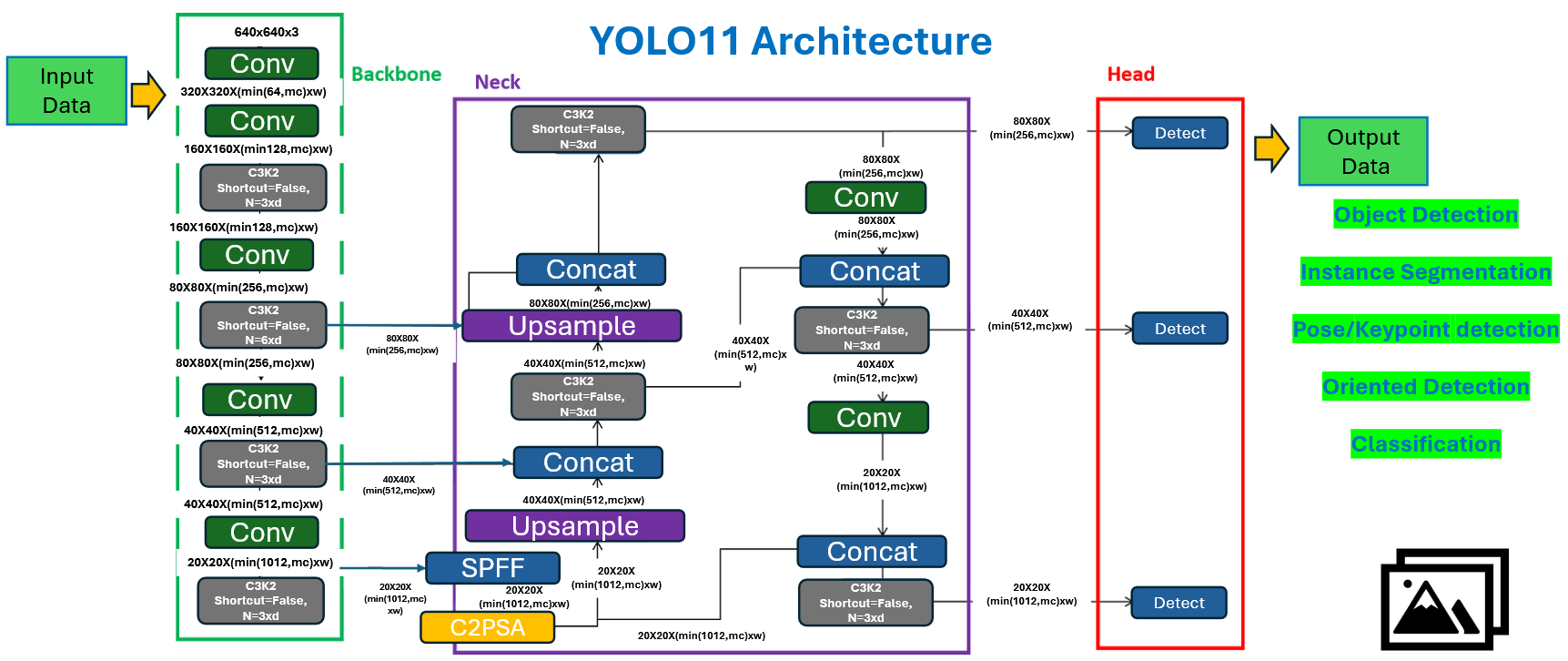}
    \footnotesize{\\Source: \cite{performanceEvalYolo}.}
    \label{fig:architectureYolo}
	\end{center}
\end{figure}

An illustration of the YOLOv11 architecture is provided in Figure \ref{fig:architectureYolo}. 

\subsection{Synthetic Dataset}
The dataset used for training and validation consisted entirely of synthetically generated images and their corresponding YOLO-format bounding box labels for a single-object class: a soup can. The base dataset, produced with Duality AI software for the Kaggle challenge \cite{kaggle}, featured only simple scenes. Recognizing the importance of data volume and diversity for sim-to-real transfer \cite{app15010354}, additional datasets from Duality AI were used. The resulting dataset consisted of 1,368 images with variations in backgrounds, camera distance, lighting, and the presence of furniture or plants.

\begin{figure}[htb!]
    \centering
    \begin{subfigure}[t]{0.23\textwidth}
        \centering 
        \includegraphics[width=\linewidth]{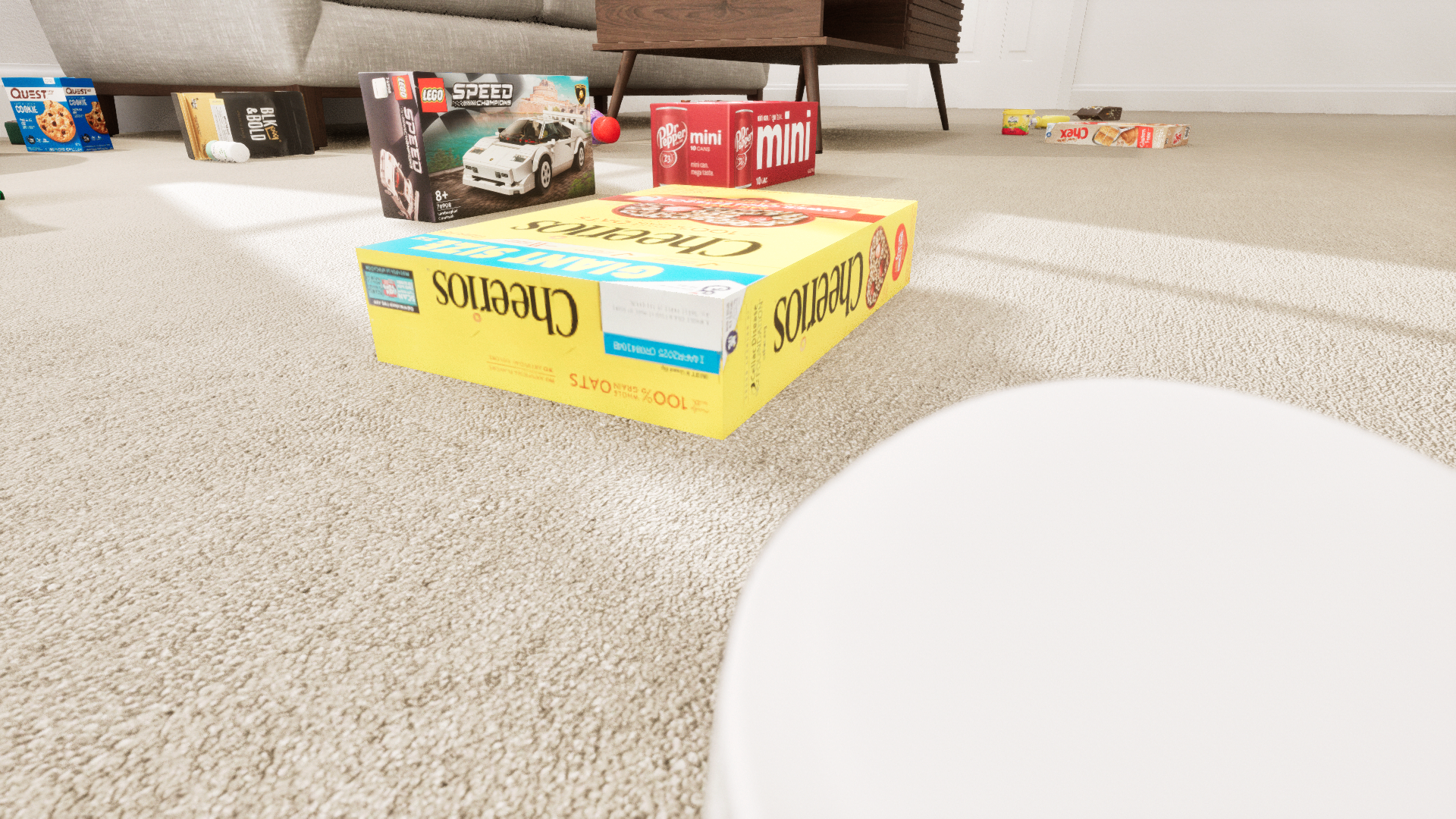}
        \caption{Synthetic image without can}
    \end{subfigure}%
    \hfill
    \begin{subfigure}[t]{0.23\textwidth} 
        \centering 
        \includegraphics[width=\linewidth]{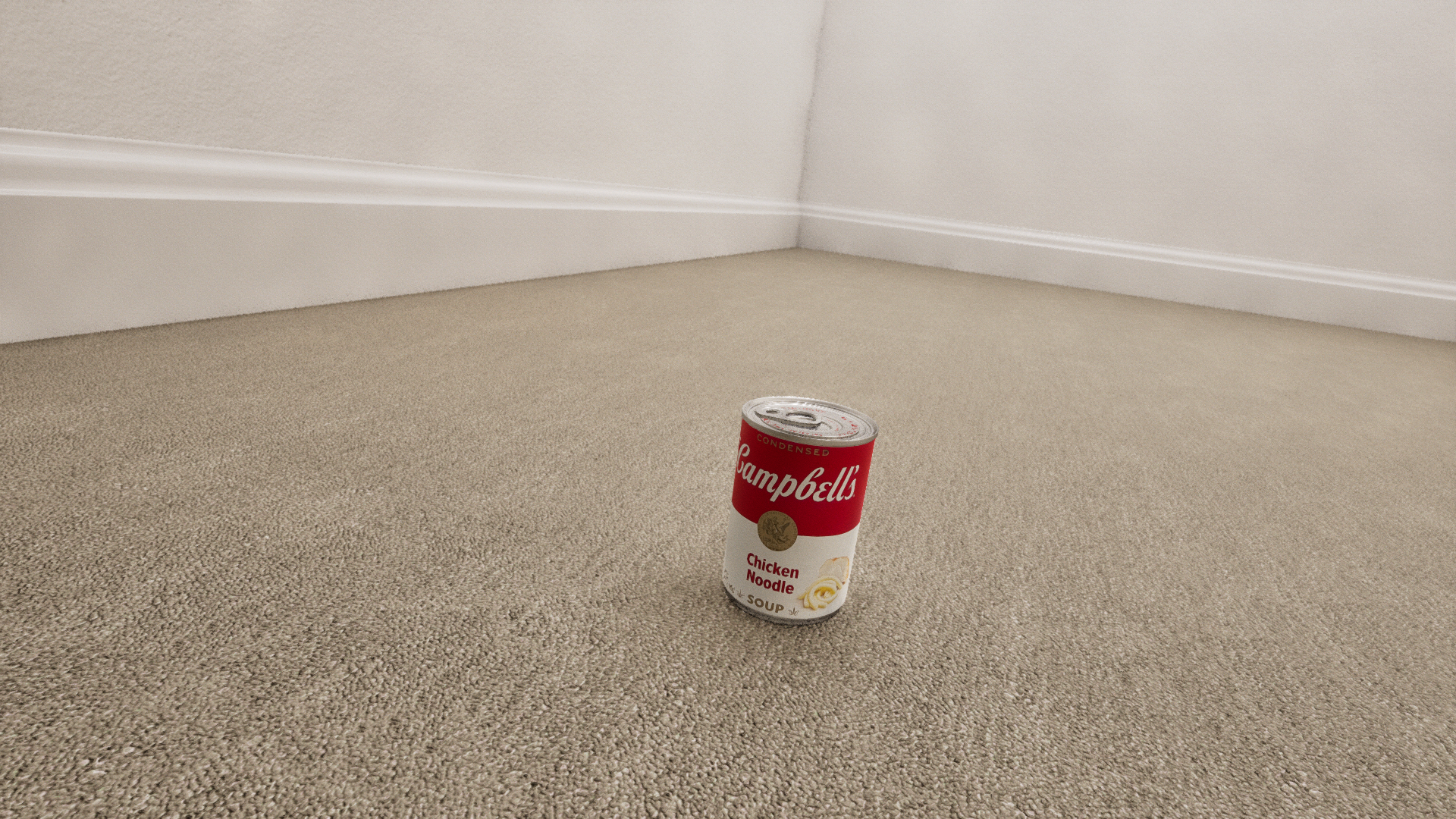} 
        \caption{Synthetic image with can}
    \end{subfigure}
    \begin{subfigure}[b]{0.23\textwidth}
        \centering
        \includegraphics[width=\linewidth]{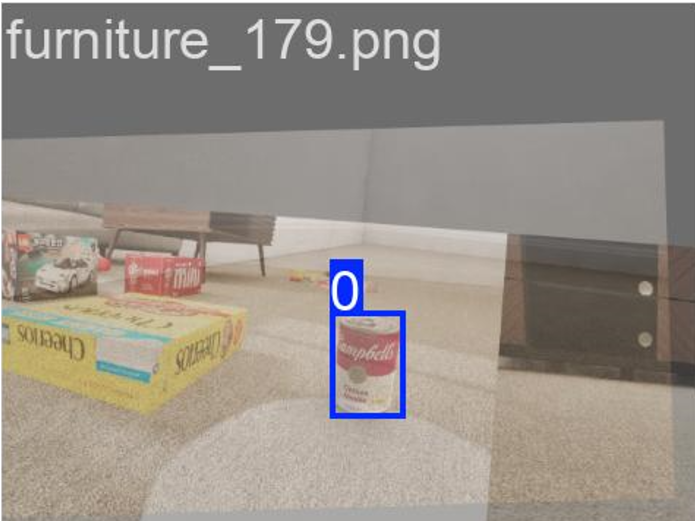}
        \caption{Mixup technique applied to images (a) and (b)}
    \end{subfigure}%
    \hfill
    \begin{subfigure}[b]{0.23\textwidth}
        \centering
        \includegraphics[width=\linewidth]{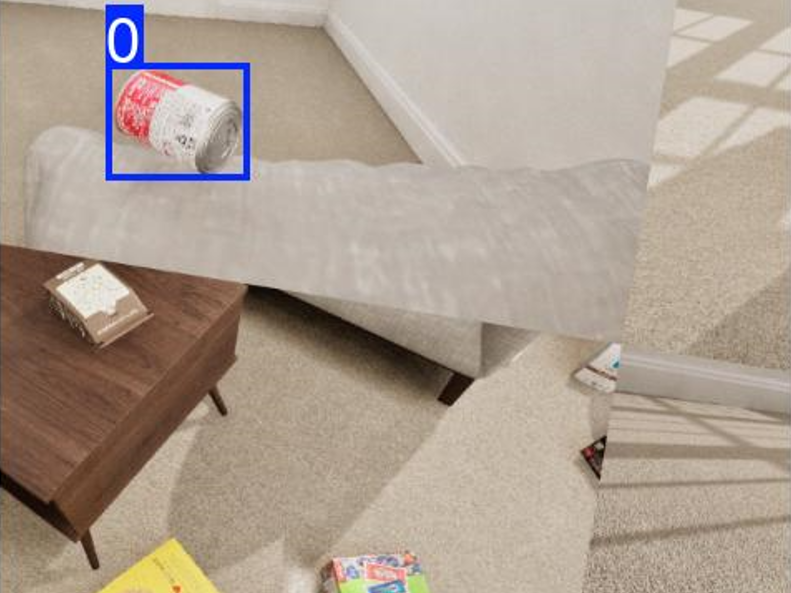} 
        \caption{Mosaic technique applied to other set of four images}
    \end{subfigure}
    \caption{Examples of (a) original complex scene without target object,
    (b) original simple scene with target object,
    (c) scene after applying mix-up augmentation to introduce a soup can image into the complex scene, (d) scene after applying mosaic augmentation combining four sample images.}
    \label{fig:data_aug_mosaic_mixup}
\end{figure}

To further increase dataset diversity and prepare the model for complex real-world scenes, including those where the target object might be present in cluttered scenes or backgrounds without the target, an additional synthetic dataset composed of 738 images containing different objects, but without soup cans, was introduced. These images included empty labels, as no bounding box targets were present. To effectively train the model to detect the target soup can even within these diverse and potentially cluttered scenes, data augmentation techniques such as Mixup and Mosaic were employed, which allowed combining images containing a soup can with images of cluttered scenes during training. Figure \ref{fig:data_aug_mosaic_mixup} illustrates the application of Mixup and Mosaic techniques. 

In total, the primary synthetic dataset used for most experiments comprised 2106 images and corresponding label files (1,368 with soup cans and 738 without). A separate set of 159 real-world images was used exclusively for testing, following the Kaggle competition guidelines \cite{kaggle}.

\subsection{Data Cleaning}
Prior to training, the integrity and quality of the synthetic dataset were verified. A data cleaning process was implemented to visually inspect each image and its associated label file. This involved developing a custom script to visualize images with their corresponding bounding boxes based on the YOLO label files. This enabled the identification and removal of images with missing label files, incorrect annotations, or other inconsistencies that could negatively impact training performance. This step was particularly important given the automated nature of synthetic data generation, which can introduce errors. 

\subsection{Training and Evaluation Strategies}
Several training and evaluation strategies were explored to develop the most robust model for sim-to-real transfer. K-Fold-Cross-Validation, a well-known technique, was implemented to obtain a more reliable estimate of model performance within the synthetic domain and to assess the robustness of different configurations. In this case, the initial five predefined dataset variations naturally served as folds. For each fold, four variations were combined for training and the remaining one was used for validation. This process was repeated for five experimental model configurations, resulting in a total of 25 trained models. This cross-validation approach aimed to provide an unbiased assessment of how different model configurations and data augmentation strategies performed across various synthetic data distributions.

Following the insights gained from the cross-validation analysis, the experiment's settings were used for training on the entire combined dataset (1.368 images) using a standard 80\% training / 20\% validation split. 

Following qualitative evaluation of predictions on the test set, the sub-dataset containing images rendered from significant distances seemed to introduce challenges or artifacts less representative of the real test set. Therefore, models were also trained on a split that excluded this particular variation to investigate its impact.

To further train a model robust to complex, cluttered real-world environments, the expanded dataset of 2106 images was used, along with advanced data augmentation methods such as Mosaic and Mixup, to add the soup can into cluttered scenes. The effectiveness of these techniques was assessed through evaluation on the test set.

Finally, acknowledging the potential benefits of model scale, the best-performing configuration found during the experiments was also trained using the medium and large architectures of YOLOv11 (YOLOv11m and YOLOv11l) in addition to the small (YOLOv11s) architecture.

\subsection{Model Configurations and Data Augmentation}\label{modelconfig}
Different model configurations were trained, primarily by varying the data augmentation applied during training. Starting from the pre-trained YOLOv11s weights, the configurations explored were:

\begin{itemize}
\item Baseline: Training with a standard set of default augmentations provided by Ultralytics' framework.
\item Augmented visual: Additional augmentations aimed at visual variability, such as perspective transformations and mixup.
\item Augmented color: Stronger color jitter augmentations to simulate wider variations in lighting and color.
\item Augmented geometry: More significant geometric transformations (rotation, translation, scale ranges) to enhance robustness to viewpoint and scale changes.
\item Augmented all: A combination of multiple augmentation strategies to leverage their potentially cumulative effects.
\end{itemize}

Validation metrics such as mAP50, mAP50-95, precision, recall, and various loss components were monitored for each configuration. Learning rates and the number of training epochs were carefully selected and tuned based on observed training dynamics and available resources.

\section{Experiments and Results}
The following experiments were conducted to evaluate the impact of different data augmentation configurations and dataset compositions on YOLOv11's performance and its generalization to real-world test images. The trained models were grouped by the dataset and strategy used for training.

\subsection{Initial 5-Fold cross-validation on the Original Synthetic Dataset}

Initial experiments involved training YOLOv11s models using a 5-fold cross-validation approach on the original synthetic dataset (1368 images), to provide an unbiased assessment of model performance across different synthetic distributions and evaluate the effectiveness of five predefined model configurations. These models were trained for 10 epochs, with batch size of 16, initial learning rate of 0.01 and final learning rate ratio of 0.01, to quickly assess performance based only on data augmentation methods variation. The average validation metrics obtained across the folds for these configurations are presented in Table \ref{ta:cross_validation}.
 
\begin{table}[htb!]
 \caption{Average metrics per configuration used in 5-fold cross validation (1368 images, 10 epochs).}
 \label{ta:cross_validation}
\resizebox{\columnwidth}{!}{%
\begin{tabular}{@{}lcccccccccccc@{}}
\toprule
\textbf{Metrics}            & \multicolumn{2}{c}{\textbf{mAP50}} & \multicolumn{2}{c}{\textbf{precision(B)}} & \multicolumn{2}{c}{\textbf{recall(B)}} & \multicolumn{2}{c}{\textbf{val/box loss}} & \multicolumn{2}{c}{\textbf{val/cls loss}} & \multicolumn{2}{c}{\textbf{val/dfl loss}} \\ \midrule
\textbf{Configuration Name} & \textbf{Avg}     & \textbf{Std}    & \textbf{Avg}        & \textbf{Std}        & \textbf{Avg}       & \textbf{Std}      & \textbf{Avg}        & \textbf{Std}        & \textbf{Avg}        & \textbf{Std}        & \textbf{Avg}        & \textbf{Std}        \\
Base                        & 0.98             & 0.01            & 0.97                & 0.02                & 0.95               & 0.04              & 0.55                & 0.16                & 0.44                & 0.09                & 0.91                & 0.07                \\
Complete                    & 0.99             & 0.02            & 0.99                & 0.03                & 0.98               & 0.04              & 1.00                & 0.08                & 0.43                & 0.08                & 0.97                & 0.05                \\
Color                       & 0.99             & 0.01            & 0.99                & 0.01                & 0.99               & 0.03              & 0.28                & 0.11                & 0.25                & 0.09                & 0.79                & 0.02                \\
Geometry                    & 0.99             & 0.00            & 0.99                & 0.01                & 0.99               & 0.02              & 0.77                & 0.37                & 0.37                & 0.07                & 0.92                & 0.13                \\
Vis. and Geo.               & 0.99             & 0.02            & 0.98                & 0.04                & 0.97               & 0.05              & 0.96                & 0.36                & 0.47                & 0.10                & 1.03                & 0.22                \\
Visual                      & 0.99             & 0.01            & 0.98                & 0.03                & 0.99               & 0.03              & 0.35                & 0.09                & 0.28                & 0.10                & 0.80                & 0.03                \\ \bottomrule
\end{tabular}
}
\end{table}

As shown in Table \ref{ta:cross_validation}, the synthetic validation performance, particularly mAP@50 was high (0.98-0.99) for all initial augmentation configurations, suggesting effective learning within the synthetic domain. However, variations in loss indicated differences in training dynamics and potential robustness. It was decided to train for more epochs and decrease the learning rate to help the model converge, with variations in the values used for the data augmentation methods as well. 

\subsection{80/20 Split Training and Qualitative Real World Evaluation}

Following the initial cross-validation, models were trained on the original synthetic dataset (1368 images) using a standard 80\% training and 20\% validation split. The hyperparameters were updated to 20 epochs and initial learning rate of 0.0001, while data augmentation values were updated based on visual observations of the model's performance on the test set. Given the challenge of bridging the domain gap, relying solely on synthetic validation metrics was insufficient. Therefore, the trained models were qualitatively evaluated by visually inspecting their predictions on the real test set, as the official labels were not available. This qualitative assessment was crucial for identifying which configurations produced fewer false positives, better localization, and more consistent detections. 

During evaluation, false positives were detected with high confidence scores, appearing similar to objects in the sub-dataset containing images rendered from significant distances. Consequently, other models were trained on a split that excluded this particular variation to investigate its impact. Table \ref{tab:splitdataset_comparison} compares models trained on a 80/20 split dataset with and without the distant camera rendering based on synthetic validation metrics. 

\begin{table}[htb!]
\caption{Comparison of models trained in 80/20 split with and without distant camera subset}
\label{tab:splitdataset_comparison}
\resizebox{\columnwidth}{!}{%
\begin{tabular}{@{}lccccccc@{}}
\toprule
\textbf{Model Name}                    & \textbf{mAP50} & \textbf{mAP50-95} & \textbf{precision(B)} & \textbf{recall(B)} & \textbf{val/box\_loss} & \textbf{val/cls\_loss} & \textbf{val/dfl\_loss} \\ \midrule
Complete                            & 1.00           & 0.97              & 1.00                  & 1.00               & 0.29                   & 0.18                   & 0.78                   \\
Vis. and Geo.                        & 1.00           & 0.94              & 1.00                  & 1.00               & 0.38                   & 0.20                   & 0.79                   \\
Without distance - Complete           & 1.00           & 0.97              & 1.00                  & 1.00               & 0.31                   & 0.18                   & 0.78                   \\
Without distance - Vis. and Geo.         & 1.00           & 0.96              & 1.00                  & 1.00               & 0.33                   & 0.18                   & 0.78                   \\ \bottomrule
\end{tabular}%
}
\end{table}

\begin{figure}[htbp]
    \centering
    \begin{subfigure}[t]{0.23\textwidth}
        \centering 
        \includegraphics[width=0.80\linewidth]{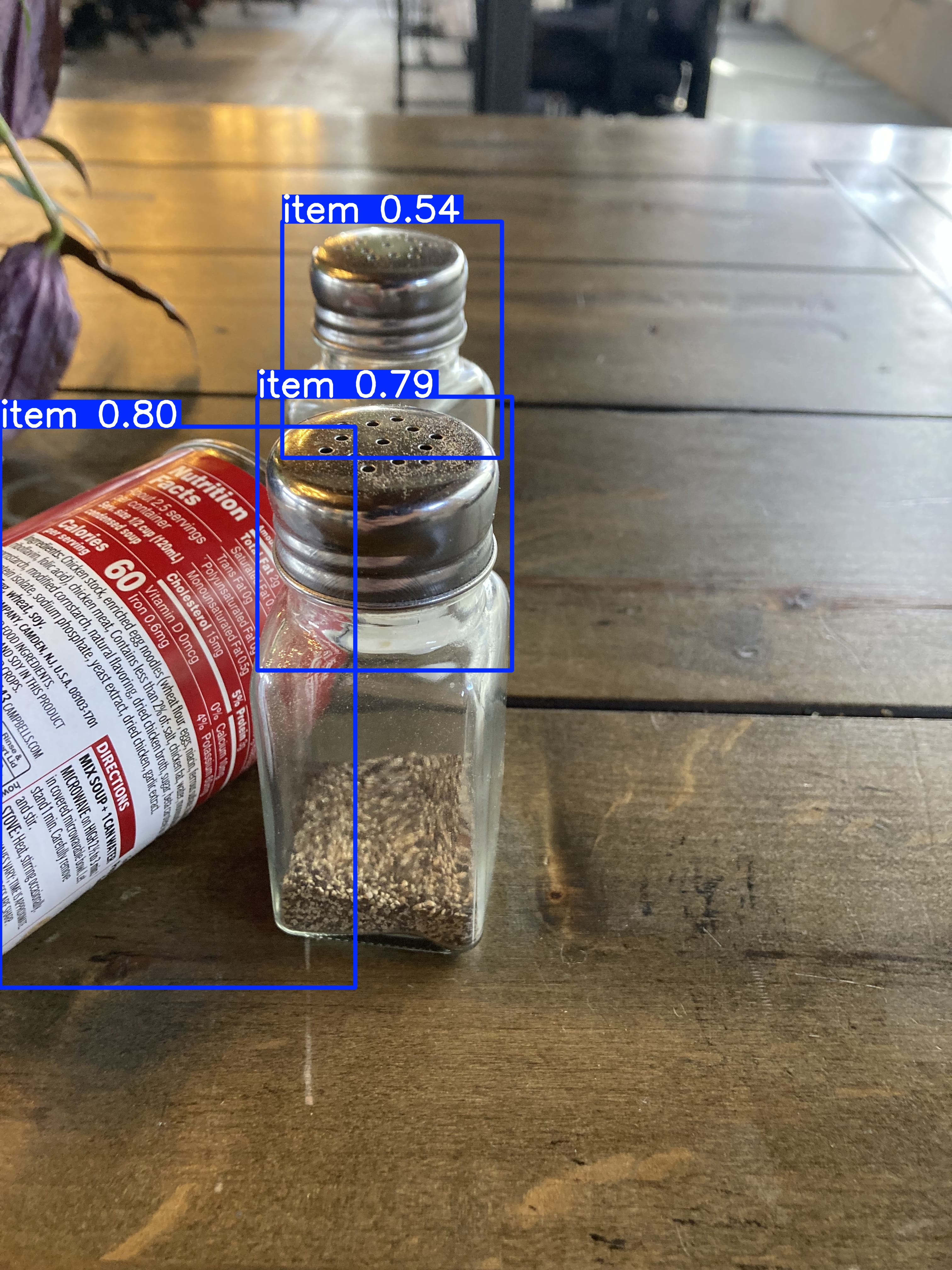}
        \caption{Prediction from model containing camera distance variant in the training dataset.}
    \end{subfigure}%
    \hfill
    \begin{subfigure}[t]{0.23\textwidth} 
        \centering 
        \includegraphics[width=0.80\linewidth]{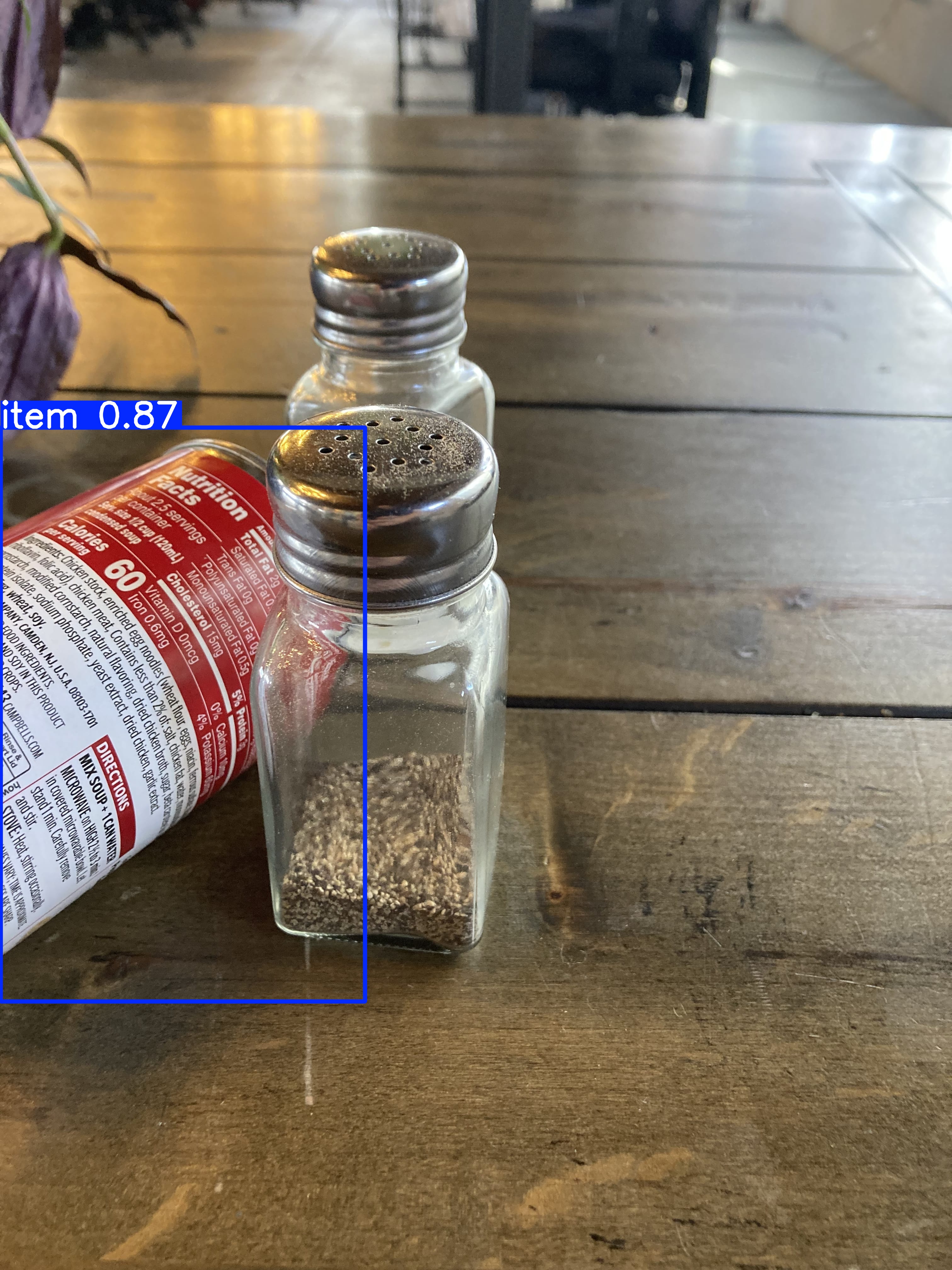} 
        \caption{Prediction from model without the camera distance variant in the training dataset.}
    \end{subfigure}
    \caption{Qualitative comparison of model predictions on a real-world test image. (a) Prediction from a model trained with the distant camera variant included in the training dataset, showing a false positive detection. (b) Prediction from a model trained with the distant camera variant excluded from the training dataset, showing the absence of the false positive.}
    \label{fig:predictions_split}
\end{figure}

While the synthetic validation metrics (Table \ref{tab:splitdataset_comparison}) show only minor differences, the decision to exclude the distant camera subset from training was driven primarily by the significant reduction in false positives observed during qualitative evaluation on real-world data. Figure \ref{fig:predictions_split} compares predictions on the same real image from models trained with and without the distant camera variant included in their training data, visually demonstrating the impact on false positives.

\subsection{5-Fold Cross-Validation on Expanded Dataset with Advanced Augmentation}

To further train a model robust to complex, cluttered environments, an expanded dataset including scenes without the target object was used with advanced data augmentation like Mosaic and Mixup. To assess the performance of different augmentation configurations on this larger dataset, a second 5-fold cross-validation was performed. The results are presented in Table \ref{tab:cv_final}.

\begin{table}[htb!]
\caption{Average metrics per configuration used in second 5-fold cross validation on expanded dataset (2106 images)}
\label{tab:cv_final}
\resizebox{\columnwidth}{!}{%
\begin{tabular}{@{}lcccccccccccc@{}}
\toprule
\textbf{Metrics}            & \multicolumn{2}{c}{\textbf{mAP50}} & \multicolumn{2}{c}{\textbf{precision(B)}} & \multicolumn{2}{c}{\textbf{recall(B)}} & \multicolumn{2}{c}{\textbf{val/box loss}} & \multicolumn{2}{c}{\textbf{val/cls loss}} & \multicolumn{2}{c}{\textbf{val/dfl loss}} \\ \midrule
\textbf{Configuration Name} & \textbf{Avg}     & \textbf{Std}    & \textbf{Avg}        & \textbf{Std}        & \textbf{Avg}       & \textbf{Std}      & \textbf{Avg}         & \textbf{Std}        & \textbf{Avg}         & \textbf{Std}        & \textbf{Avg}         & \textbf{Std}        \\
Complete                    & 0.99             & 0.00            & 0.99                & 0.01                & 0.99               & 0.01              & 0.33                 & 0.12                & 0.24                 & 0.12                & 0.78                 & 0.01                \\
Base                        & 0.99             & 0.00            & 0.99                & 0.01                & 0.99               & 0.01              & 0.20                 & 0.10                & 0.20                 & 0.14                & 0.77                 & 0.01                \\
Color                       & 0.99             & 0.00            & 0.99                & 0.01                & 0.98               & 0.02              & 0.20                 & 0.11                & 0.22                 & 0.20                & 0.77                 & 0.01                \\
Geometry                    & 0.99             & 0.01            & 0.98                & 0.04                & 0.99               & 0.02              & 0.51                 & 0.07                & 0.31                 & 0.15                & 0.82                 & 0.03                \\
Vis. and Geo.               & 0.99             & 0.00            & 1.00                & 0.01                & 0.99               & 0.01              & 0.33                 & 0.17                & 0.23                 & 0.13                & 0.78                 & 0.01                \\ \bottomrule
\end{tabular}%
}
\end{table}

The results in Table \ref{tab:cv_final} show consistently high mAP@50 across the various augmentation configurations on the expanded dataset, similar to the initial CV. A comparison of Table \ref{ta:cross_validation} (original dataset, 10 epochs) and Table \ref{tab:cv_final} (expanded dataset, 20 epochs, different split strategy) shows that mAP@50 remained high, while the overall loss decreased significantly, suggesting a positive result with an expanded dataset and longer training implementation. 

\subsection{Training on Expanded Dataset with Advanced Augmentation}

Models were then trained using the two data augmentation configurations that achieved the best results during experimentation, hereafter referred to as "A" and "B" for simplicity. Configuration A represents a comprehensive augmentation strategy, including geometric (rotation, translation, scale), color space (HSV), and compositional methods (mosaic, mixup), while Configuration B used the same techniques, but without mosaic and with different parameter values.

To evaluate the impact of all previously proposed improvements, they were trained on split datasets with and without camera distance, and then with the addition of the expanded dataset without soup cans. Recognizing the potential benefits of model scale, all variations were also trained on large (YOLOv11l) architectures. To estimate the mAP@50 on the hidden real test set, the test set images were manually annotated, providing valuable initial insight into the models' performance. Table \ref{tab:map50_priorcalculation} summarizes the best-performing models across different training dataset compositions, model scales, and data augmentation configurations. It includes mAP@50 calculated with the synthetic validation set together with the mAP@50 score obtained on the manually annotated real test set, illustrating the correlation between synthetic and real-world performance.

\begin{table}[htb!]
\caption{Comparison of best performing models across dataset compositions, model scales and data augmentation configurations according to mAP@50 calculated on real data test set}
\label{tab:map50_priorcalculation}
\resizebox{\columnwidth}{!}{%
\begin{tabular}{@{}lcccc@{}}
\toprule
\multicolumn{1}{c}{\textbf{Dataset Compositions}}                           & \textbf{Model Scale} & \textbf{\begin{tabular}[c]{@{}c@{}}Data Augmentation\\ Configuration\end{tabular}} & \textbf{\begin{tabular}[c]{@{}c@{}}map@50 \\ validation set\end{tabular}} & \textbf{\begin{tabular}[c]{@{}c@{}}map@50 \\ test set\end{tabular}} \\ \midrule
Standard Split                                                              & Large                & A                                                                                  & 0.995                                                                     & 0.877                                                               \\
Split without distance                                                      & Large                & A                                                                                  & 0.995                                                                     & 0.885                                                               \\
\begin{tabular}[c]{@{}l@{}}Expanded dataset\\ With distance\end{tabular}    & Large                & A                                                                                  & 0.995                                                                     & 0.952                                                               \\
\begin{tabular}[c]{@{}l@{}}Expanded dataset\\ Without distance\end{tabular} & Large                & A                                                                                  & 0.995                                                                     & 0.930                                                               \\ \bottomrule
\end{tabular}%
}
\end{table}

As expected, the mAP@50 score from the synthetic validation set diverged significantly from the score on the real-world test set, indicating the domain gap. Notably, the configuration that achieved the highest performance was A, which applies all available data augmentation methods from Ultralytics. As expected, the use of large models and the expanded dataset was beneficial to performance. The models and corresponding predictions that achieved a high mAP@50 on the manually annotated test set were selected for formal submission to the competition platform. The final mAP@50 score was calculated by the organizers using the predicted bounding boxes and confidence scores. Table \ref{tab:map50_competitioncalculation} summarizes the real-world test set mAP@50 scores reported by the competition platform, compared to those previously calculated on the hand-labeled test set.

\begin{table}[htb!]
\caption{Comparison of best performing models with mAP@50 calculated by competition organization}
\label{tab:map50_competitioncalculation}
\resizebox{\columnwidth}{!}{%
\begin{tabular}{lcccc}
\hline
\multicolumn{1}{c}{\textbf{Dataset Compositions}}                           & \textbf{Model Scale} & \textbf{\begin{tabular}[c]{@{}c@{}}Data Augmentation\\ Configuration\end{tabular}} & \textbf{\begin{tabular}[c]{@{}c@{}}map@50 \\ test set\end{tabular}} & \textbf{\begin{tabular}[c]{@{}c@{}}map@50\\ competition\end{tabular}} \\ \hline
Standard Split                                                              & Large                & A                                                                                  & 0.877                                                               & 0.905                                                                 \\
Split without distance                                                      & Large                & A                                                                                  & 0.885                                                               & 0.885                                                                 \\
\begin{tabular}[c]{@{}l@{}}Expanded dataset\\ With distance\end{tabular}    & Large                & A                                                                                  & 0.952                                                               & 0.910                                                                 \\
\begin{tabular}[c]{@{}l@{}}Expanded dataset\\ Without distance\end{tabular} & Large                & A                                                                                  & 0.930                                                               & 0.900                                                                 \\ \hline
\end{tabular}%
}
\end{table}

Before evaluating the results, it is important to understand the primary metric of the challenge. mAP@50 stands for Mean Average Precision with an Intersection over Union (IoU) threshold of 50\%. Precision measures how accurate the model's positive predictions are, while IoU determines whether a detected bounding box is a true positive or false positive. If the overlap between a predicted bounding box and the ground truth is greater than 50\%, the prediction is considered a true positive, otherwise it is a false positive.

The best real-world test set mAP@50 score was 0.910. This was achieved by the YOLOv11l model trained on the split dataset with camera distance and addition of expanded dataset, using configuration A for data augmentation. The addition of an extended dataset positively contributed to predictions in a complex real test environment with multiple objects from other classes. Although it was thought that excluding the camera distance variant would contribute positively to the model, given the constrained and small size of the available dataset, feeding more varied perspectives of the object proved to be crucial. 

It is important to note the difference between the mAP@50 score calculated on our manually annotated labels and the official score provided by the competition. The first one was used as an estimation tool to overcome the challenge of evaluating the sim-to-real gap, as validation alone did not contribute to the analysis of real-world performance. Moreover, the methods used for mAP@50 calculation differed from those of the competition, whose methodology was not publicly available.  

\begin{figure}[htb!]
    \begin{center}
    \caption{Training and validation loss and metric curves for the final best-performing YOLOv11l model trained on the expanded dataset.}
    \includegraphics[width=8cm]{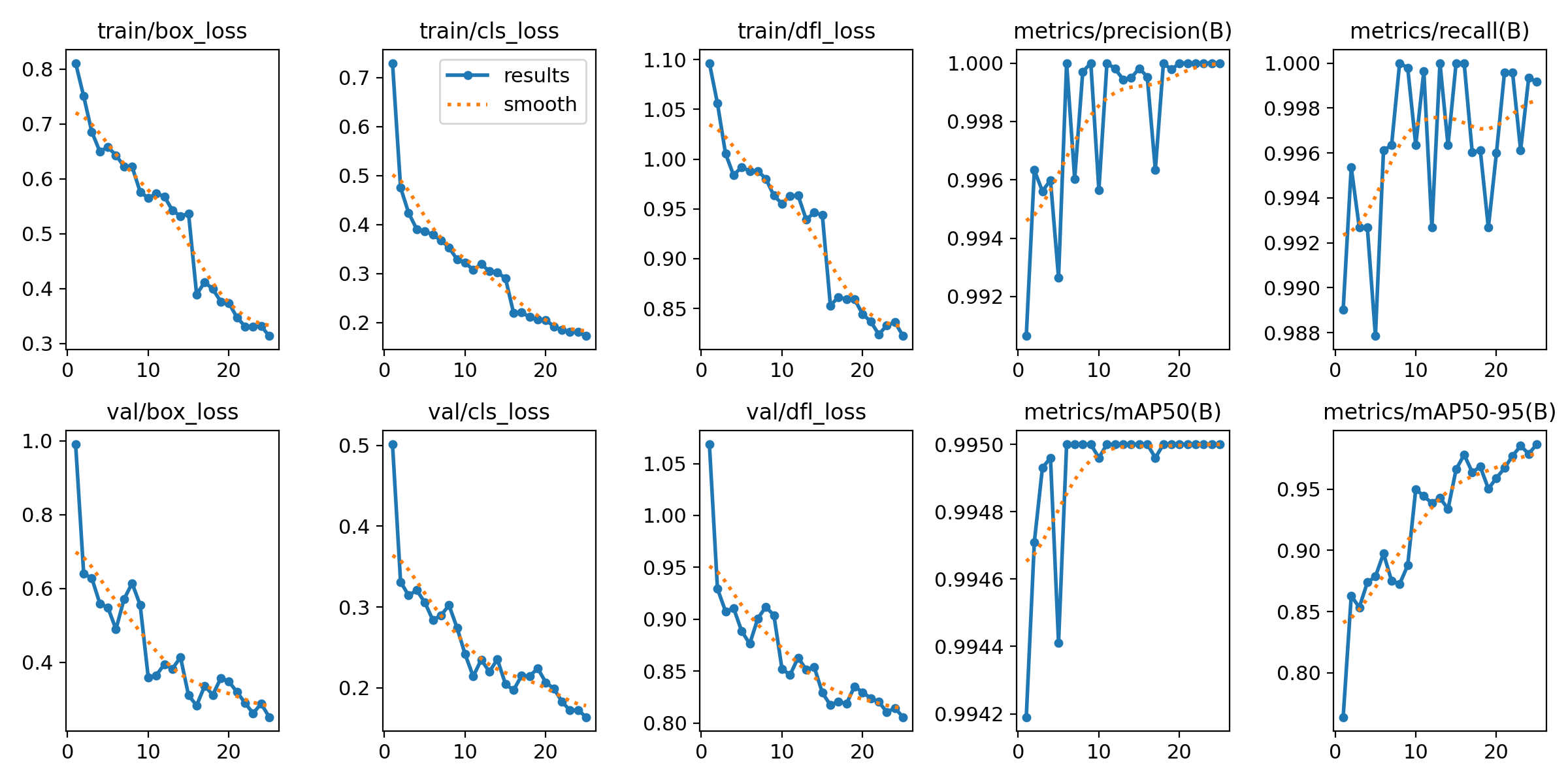}
    \label{fig:yolo11l_expanded_withdis_aug_all7_results}
	\end{center}
\end{figure}

Figure \ref{fig:yolo11l_expanded_withdis_aug_all7_results} illustrates the training and validation metric curves for the YOLOv11l trained on the expanded dataset with a strong augmentation strategy. The curves show a decrease in loss and an increase in metrics such as mAP@50 as training progresses, indicating model convergence and improved performance on the validation set.

\begin{figure}[htb!]
    \begin{center}
    \caption{Batch of predicted labels by the final best-performing model on diverse real-world test images.}
    \includegraphics[width=7cm]{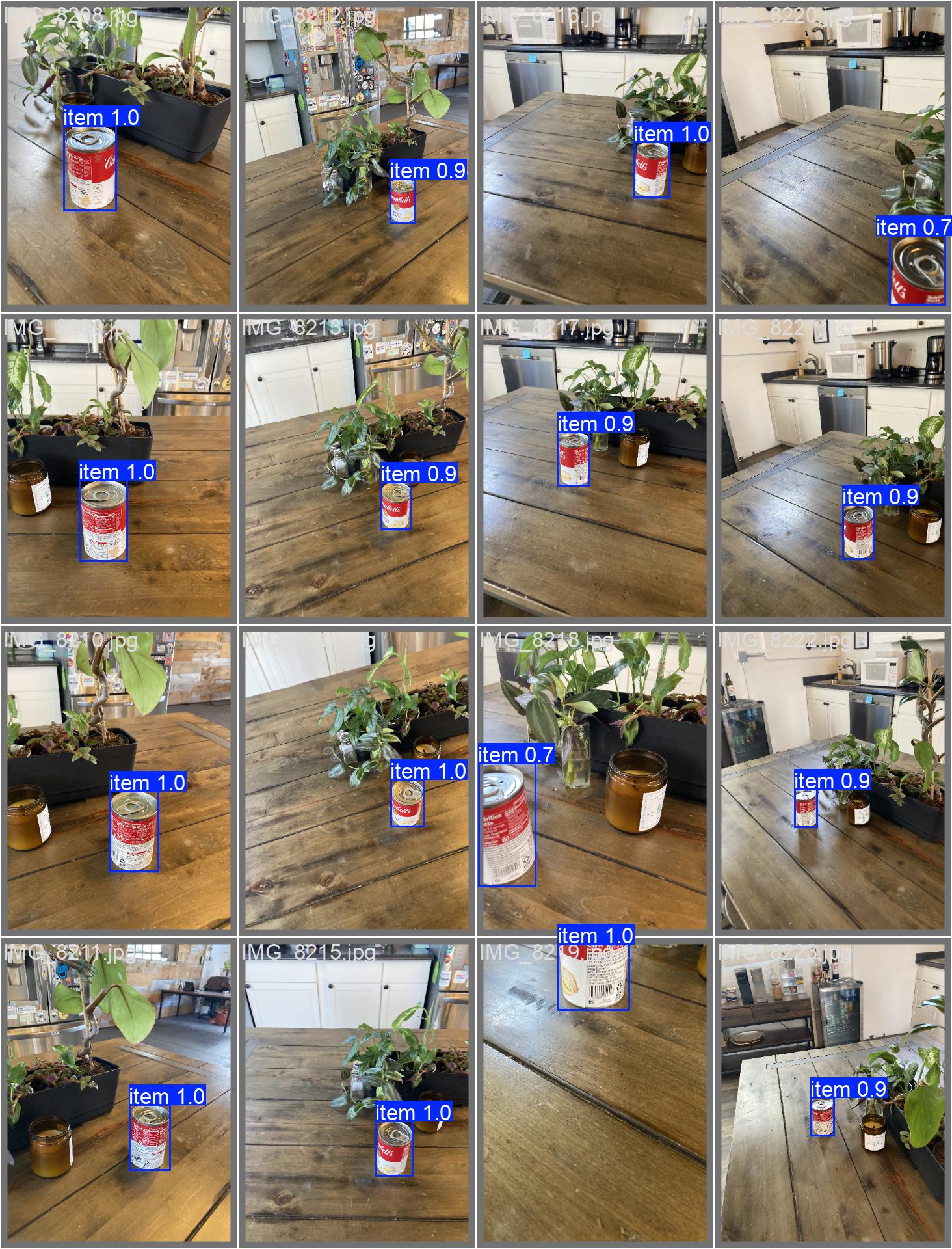}
    \label{fig:predictions}
	\end{center}
\end{figure}

Figure \ref{fig:predictions} presents visual examples of successful detections and common failure cases, such as low confidence scores when the object is only partially visible, from the final best-performing model on the real test set, providing insights into its performance characteristics.

\section{Discussion}
This project aimed to address the synthetic-to-real domain gap for object detection by training a YOLOv11 model solely on synthetic data enhanced with domain randomization strategies, primarily through data augmentation, within the context of a Kaggle competition \cite{kaggle}.

From the average cross-validation performance on the synthetic dataset (Table \ref{ta:cross_validation}), it was evident that the synthetic validation mAP@50 scores were consistently high across all the data augmentation strategies explored (0.98-0.99). This indicated that the model was capable of achieving high accuracy within the synthetic domain, regardless of the specific augmentation scheme. Despite high synthetic validation scores, the relationship between synthetic validation performance and actual performance on real data was complex. High performance on synthetic validation data alone is not a sufficient indicator of performance on real-world data due to the inherent domain gap. The real test set presented a complex environment with variations that were not perfectly captured by the synthetic data.

Moreover, variations in the loss metrics suggested that different augmentations and hyperparameters influenced the training dynamics and potentially affected generalization. The overall trend showed improved convergence stability with longer training times and smaller learning rates, while the effects of domain randomization had to be analyzed further. 

To address the limitations of the initial synthetic dataset, an iterative refinement strategy was adopted to increase scene complexity and visual variability. This involved, among other techniques, expanding the dataset with images lacking the target object and using compositional augmentations, like Mixup and Mosaic, to create more varied training examples. The effectiveness of these techniques was determined empirically: models were trained, and their predictions on the real-world test set were analyzed to guide each subsequent iteration. This feedback loop, conducted without official test set labels, proved crucial for tuning the domain randomization strategy. For example, this process revealed that configurations using Mixup and Mosaic on the expanded dataset demonstrated superior robustness in cluttered scenes. In contrast, it also showed that overly aggressive geometric augmentations could lead to mislocalization, helping to find a balance that improved detection accuracy.

Furthermore, visual inspection of the predictions on the test set initially suggested that synthetic images rendered from a distant perspective were harmful to model generalization, contributing to false positives (see Figure \ref{fig:predictions_split}). Based on this observation, this dataset variation was initially excluded from some training runs to investigate its impact. However, the final quantitative results on the real test set revealed that the model trained on the expanded dataset with the distant viewpoint achieved the highest mAP@50 score calculated by the competition. This suggests that while potentially introducing some initial false positives, including varied perspectives of the object in the training data ultimately contributed positively to the model's ability to generalize to a broader range of real-world scenarios, likely by increasing the overall diversity of the training set.

The configuration that achieved the highest mAP@50 on the real test set (0.910) was one that combined an expanded and diverse synthetic dataset with a carefully tuned set of augmentation strategies, demonstrating the effectiveness of Domain Randomization strategies for sim-to-real tasks. Although many advanced methods use complex techniques such as adversarial domain adaptation \cite{ganin2015unsuperviseddomainadaptationbackpropagation} \cite{tzeng2017adversarialdiscriminativedomainadaptation}, or GAN-based image-to-image translation \cite{8237506}, the results presented here prove that a simpler approach can be just as effective for specific problems. It was found that strategically expanding the diversity of the synthetic dataset, by adding varied viewpoints and negative examples, had a greater impact than any augmentation technique alone. This suggests that focusing on the quality and variety of synthetic data can achieve good real-world performance, avoiding complex model architectures or difficult training procedures.

However, it also highlights the remaining challenges. Factors not adequately captured by the synthetic data likely contributed to false negatives or positives on the real test set. Future strategies for sim-to-real transfer could benefit from even more realistic or diverse synthetic data generation, with cluttered scenes and occlusions, incorporating advanced domain adaptation techniques, or exploring methods to reduce the discrepancy between synthetic and real data distributions. 

\section{Conclusion}
This project investigated the feasibility of training a robust object detection model for a specific object (soup can) using only synthetic data and domain randomization strategies (primarily data augmentation) to achieve high performance on unseen real-world images. Using the YOLOv11 architecture, various data augmentation configurations were explored and evaluated, initially through cross-validation on the synthetic dataset and subsequently through qualitative assessment on a real-world test set.

While synthetic validation metrics consistently showed high mAP@50 across different augmentation strategies, it was found that these metrics alone were insufficient indicators of real-world performance due to the sim-to-real domain gap. Therefore, the results were evaluated qualitatively, through visual inspection, and later quantitatively using manually annotated labels, given that the official labels were reserved to the competition organizers, which imposed an additional challenge to assess model performance on real-world data. 

The key to improving real-world generalization lay in expanding the synthetic dataset to include more complex backgrounds and carefully tuning the data augmentation parameters based on empirical observation and qualitative analysis of predictions on real-world images.

The configuration that produced the highest performance on the real test set, achieving a mAP@50 of 0.910, was the YOLOv11l model trained on the expanded synthetic dataset with a carefully tuned combination of visual, geometric, and compositional (Mixup/Mosaic) augmentations. This demonstrates that domain randomization via well-tuned data augmentation can be an effective strategy for bridging the synthetic-to-real gap, even without access to real-world training data.

However, the achieved score also highlights the remaining challenges in fully eliminating the domain gap using this approach alone. Future work could explore the generation of synthetic data that better captures the full complexity and variability of the real-world or investigate more advanced domain adaptation techniques.

In real-world applications, synthetic data is often combined with real-world data to help bridge the sim-to-real transfer gap. The primary challenge imposed by this competition was to accomplish this using only synthetic data.

\section{References}


\end{document}